\documentclass[preprint, 5p]{elsarticle}
\usepackage{hyperref}
\usepackage{float}
\usepackage{verbatim} 
\usepackage{amsmath}
\restylefloat{figure}
\restylefloat{table}
\usepackage{lineno}

\journal{Journal of Supercomputing}

\begin{document}
\begin{frontmatter}

\title{Exploring Multi-Agent Reinforcement Learning for Unrelated Parallel Machine Scheduling}

\author[vicom,nap]{Maria Zampella \corref{cor1}}
\ead{maria.zampella2@studenti.unina.it}

\author[vicom,ehu]{Urtzi Otamendi}
\ead{uotamendi@vicomtech.org}

\author[vicom]{Xabier Belaunzaran}
\author[vicom,ehu]{Arkaitz Artetxe}
\author[vicom]{Igor G. Olaizola}
\author[nap]{Giuseppe Longo}
\author[ehu]{Basilio Sierra}

\cortext[cor1]{Corresponding author.}
\address[vicom]{Vicomtech Foundation, Basque Research and Technology Alliance (BRTA), Donostia-San Sebastián 20009, Spain}
\address[ehu]{Department of Computer Sciences and Artificial Intelligence, University of the Basque Country (UPV/EHU), Donostia-San Sebastián 20018, Spain}
\address[nap]{ Department of Physics "Ettore Pancini", University of Naples "Federico II", Naples 80126, Italy }

\begin{abstract}
Scheduling problems pose significant challenges in resource, industry, and operational management. This paper addresses the Unrelated Parallel Machine Scheduling Problem (UPMS) with setup times and resources using a Multi-Agent Reinforcement Learning (MARL) approach. The study introduces the Reinforcement Learning environment and conducts empirical analyses, comparing MARL with Single-Agent algorithms. The experiments employ various deep neural network policies for single- and Multi-Agent approaches. Results demonstrate the efficacy of the Maskable extension of the Proximal Policy Optimization (PPO) algorithm in Single-Agent scenarios and the Multi-Agent PPO algorithm in Multi-Agent setups. While Single-Agent algorithms perform adequately in reduced scenarios, Multi-Agent approaches reveal challenges in cooperative learning but a scalable capacity. This research contributes insights into applying MARL techniques to scheduling optimization, emphasizing the need for algorithmic sophistication balanced with scalability for intelligent scheduling solutions.
\end{abstract}


\end{frontmatter}

\section{Introduction}
\label{introduction}

Scheduling problems constitute a subset of optimization challenges that find widespread applications across various sectors, encompassing resource management \citep{ravindran2005recent}, industry \citep{jiang2022evolution, zhang2016planning}, and operational management \citep{graves1981review}. In particular, production scheduling, an essential facet of manufacturing, revolves around the efficient and cost-effective allocation of limited resources to support production processes. In this context, implementing flexible and intelligent strategies for industrial scheduling emerges as an imperative.

The primary objective of scheduling problem optimization is to identify an advantageous combination of decision variables within a defined search space. These decision variables dictate the order in which processes or tasks are assigned to a set of machines, typically entailing complex combinatorial problems. These problems often involve optimizing multiple objectives, depending on the system demands \citep{nagar1995multiple}. In industrial settings, the overall aim predominantly implies minimizing job completion times while accommodating other objectives, such as resource utilization or environmental considerations.

It is important to note that real-world industrial scheduling problems imply exploring extensive search spaces, usually culminating in NP-hard problem instances  \citep{lenstra1977complexity}. This inherent intricacy poses complex challenges to the research and development community. This characteristic has attracted considerable attention from the research community, as exemplified in the comprehensive survey by Allahverdi et al. \cite{allahverdi2008survey}.

Notwithstanding the advancements in this field, current approaches have notable limitations \citep{abdullah2014fuzzy}, encompassing computational complexity \citep{ren2022joint} and the capacity to generalize and adapt to diverse problem instances \citep{yan2021research, zhang2020improved}. In response to these constraints, contemporary research has increasingly employed advanced decision-making techniques.

Deep Learning techniques have obtained significant attention in intelligent decision-making systems within the research community \citep{liu2023integrating}. Specifically, Reinforcement Learning (RL) approaches have emerged as valuable solutions to addressing scheduling in complex environments \citep{wang2021review}. This machine learning paradigm relies on utilizing an intelligent agent that learns from the actions it takes and the rewards it receives in response to those actions. Reinforcement Learning models exhibit adaptability to non-deterministic environments, making it a flexible approach for optimization within dynamic and uncertain environments.

This paper presents a study of a Multi-Agent Reinforcement Learning (MARL) approach to addressing an optimal job scheduling problem. The research is performed in the Unrelated Parallel Machine scheduling problem (UPMS) with setup times and resources proposed by Fanjul et al. \cite{fanjul2020models}, a single-stage job scheduling problem variant. 

This research presents the novelty of employing a MARL approach that extends beyond traditional methods. In this sense, the work reviews the RL environment deployed in the study and conducts empirical analyses, comparing the MARL approach with various Single-Agent algorithms. The contribution aims to provide a practical evaluation of the efficacy of addressing complex decision-making problems like UPMS.

The remainder of the article is structured as follows:
Section \ref{sec: related_work} provides an overview of reinforcement learning on scheduling problems.
Section \ref{sec: problem} formally describes the problem addressed in the paper. 
Section \ref{sec: environment} describes the implementation details. 
Section \ref{sec: methodology} introduces the proposed approach and implementation insights. 
Section \ref{sec: Experiments} presents the evaluation and validation process conducted in the experiments. Finally, section \ref{sec: conclusions} contains the concluding remarks and outlines some areas for further research.

\section{Related work}
\label{sec: related_work}

\textit{Job scheduling} is a fundamental optimization process used in various industries \citep{zhang2019review}. This field involves task allocation to machines to optimize objectives. This optimization class encompasses diverse variations based on task types, machines, and constraints \citep{graham1979optimization}, classifying them into single- and multi-step processes.

The job scheduling problem is a well-known intricate combinatorial optimization issue within machine scheduling, long identified as NP-hard \citep{lenstra1977complexity}. Due to its formidable complexity, obtaining an optimal solution in a reasonable timeframe is frequently unattainable \citep{asadzadeh2010agent}. The literature has proposed diverse approaches to tackle this challenge, including neural networks and deep learning methods. Genetic algorithms (GAs), initially introduced by Holland et al. \cite{holland1992adaptation}, are actively explored, yielding promising results \citep{pezzella2008genetic,kacem2002approach}. Evolutionary Algorithm (EA) counterparts such as Differential Evolution (DE) \citep{xu2013hybrid} have also received attention. Ant Colony Optimization \citep{colorni1992investigation}, as exemplified by \citep{rossi2007flexible} in Flexible Job Shop Scheduling (FJSS), optimizes setup and transportation time. 

Similarly, Artificial Bee Colony techniques by Karaboga et al. \cite{karaboga2005idea} and improved approaches by Lei et al. \cite{lei2021improved} have shown promising results in multi-objective problems for distributed parallel machine scheduling. Greedy randomized adaptive search (GRASP) \citep{feo1989probabilistic}, a widely-used metaheuristic algorithm, finds application in minimizing makespan and workloads under resource constraints \citep{rajkumar2010grasp}. Particle swarm optimization (PSO) \citep{eberhart1995new}, inspired by collective behaviour in nature, has also seen success in Job Shop Scheduling (JSS) problems \citep{sadrzadeh2013development}. Simulated Annealing (SA) \citep{kirkpatrick1983optimization} has been applied effectively, notably in makespan minimization by Loukil et al. \cite{loukil2007multi} and job-machine assignment and sequencing by Yazdani et al. \cite{yazdani2015two}. Recent decades have witnessed the adoption of heuristics and metaheuristics to address scheduling optimization challenges, with hybrid approaches integrating optimization techniques gaining prominence \citep{pellerin2019review}.

In this sense, researchers have devised mathematical models to optimize diverse objectives. He et al. \cite{he2013scheduling} introduce a novel model for the FJSS problem, considering machine breakdowns through non-cooperative game theory. Demir et al. \cite{demir2013evaluation} explore various mathematical formulations for FJSS, offering a time-indexed model for makespan minimization. Concerning the Unrelated Parallel Machine Scheduling (UPMS) problem, exact methods like Mixed-Integer Linear Programming (MILP) have been proposed, adeptly managing complex constraints and delivering optimal solutions for relatively small instances \citep{fanjul2017models, fanjul2019reformulations}.

These methodologies are categorized as static scheduling methods, which make decisions at compilation time and rely on prior task knowledge. In this sense, static scheduling techniques necessitate complete and precise task information, which is challenging to obtain in uncertain environments. In contrast, dynamic scheduling, operating during execution, utilizes computational state information for decision-making.

Reinforcement Learning (RL) has emerged as an advantageous approach for dynamic task scheduling due to its capacity to handle environmental uncertainty in dynamic contexts, its ability to learn from experience, computational efficiency, and high adaptability. Reyna et al. \cite{reyna2015reinforcement} presented the Q-Learning Reinforcement Learning algorithm to solve Job Shop and Flow Shop scenarios. In this work, the authors validate the approach's efficacy by comparing results against reported optimal solutions in specialized literature. Extending the mentioned approach, \citep{oren2021solo} employing Deep Q-Learning with graph representations of states accommodating dynamic problem potential deviations by incorporating graph-convolutional policies into Monte Carlo Tree Search. 
Li et al. \cite{li2023two} proposed a two-stage RNN-based deep reinforcement learning approach, demonstrating strong generalization capabilities in extensive numerical experiments. However, these works need more exploration into the potential sensitivity of the proposed agent to changes in problem characteristics.
To overcome the limitation of the Single-Agent approach, Liu et al.\cite{LIU2023106294} propose a deep Multi-Agent reinforcement learning-based approach to solve the dynamic job shop scheduling problem by using a centralized training and decentralized execution scheme and parameter-sharing technique to tackle the non-stationary problem. 

Despite the extensive literature on scheduling methodologies, potential benefits and challenges associated with integrating Multi-Agent techniques in this domain still need to be explored. Our research addresses this gap by investigating and proposing innovative Multi-Agent solutions for scheduling problems.

\section{Problem description}
\label{sec: problem}

In this section, we formally define the problem of Unrelated Parallel Machine Scheduling (UPMS) with resources and setup time and introduce the relevant mathematical notation. This problem focuses on efficiently allocating jobs to a group of Unrelated Parallel Machines, with due consideration given to setup times, resource utilization, and human-operator capabilities. Additionally, we present an illustrative example.

\subsection{Mathematical Notation}

The mathematical notation of UPMS in this paper is as follows:

\begin{itemize}
    \item \textbf{Machines}: a set of $M$ unrelated parallel machines, denoted as $\mathcal{M} = \{m_i \mid i \in \{1, 2, \ldots, M\}\}$.
    \item \textbf{Jobs}: a set of $J$ jobs, denoted as $\mathcal{J} = \{j_i \mid i \in \{1, 2, \ldots, J\}\}$.
    \item \textbf{Processing Time}: the time required to process job $j$ on machine $m$, denoted as $pt_{jm}$.
    \item \textbf{Setup Time}: the time required to transition time from job $j_i$ to job $j_k$ on machine $m$, denoted as $st_{j_{i}j_{k}m}$.
\end{itemize}

The worker concept is introduced to address the resource utilization:

\begin{itemize}
    \item \textbf{Workers}: A set of $W$ workers, denoted as $\mathcal{W} = \{w_i \mid i \in \{1, 2, \ldots, W\}\}$.
    \item \textbf{Worker-Machine Compatibility}: Binary variable $o_{wm}$ represents whether worker $w$ can operate machine $m$, where $o_{wm} = 1$ implies compatibility and $o_{wm} = 0$ otherwise.
    \item \textbf{Required Workforce}: The number of workers needed to perform job $j$ on machine $m$, denoted as $r_{jm}$.
\end{itemize}

The primary objective of the UPMS problem is to efficiently allocate tasks to machines and determine the sequence in which these tasks are executed on each machine. This considers various parameters, including processing times, setup times, or workforce constraints. 

The optimization objectives are dependent on the requirements and intended behaviour. In this example of the UPMS problem, a multi-objective function is employed to assess solution optimality. This function integrates diverse objectives, including minimizing overall task completion time, reducing setup times, and optimizing resource utilization. The primary challenge involves proficiently allocating tasks across machines and scheduling them to achieve these objectives while respecting constraints associated with worker-machine compatibility.

\begin{equation}
\text{Min } f(\mathbf{x}) = w_1 \cdot T(\mathbf{x}) + w_2 \cdot U(\mathbf{x})  - w_3 \cdot P(\mathbf{x})
\label{eq: multi-objective}
\end{equation}

Where:
\begin{align*}
    & T(\mathbf{x}) \text{ represents the overall task completion time}, \\
    & \text{considering job processing times and setup times}, \\
    & U(\mathbf{x}) \text{ represents the resource utilization}. \\
    & P(\mathbf{x}) \text{ represents the number of jobs performed}.
\end{align*}

The weights $w_1$, $w_2$, and $w_3$ are used to adjust the importance of each objective based on the optimization priorities within the multi-objective function. These weights can be set differently depending on the specific goals of the manufacturing plant. For example, if the goal is to reduce resource usage, the model has to decide to assign jobs to machines with lower resource requirements, even if this results in an increase in processing time.

\subsection{Illustrative Example}

We present a practical scenario within a manufacturing context is considered. In this scenario, three machines (designated as $M=3$) are considered for processing five jobs ($J=5$). The workforce consists of two workers ($W=2$). The analysis involves the utilization of processing times ($pt_{jm}$, as introduced in Table \ref{tab:example_pt}), setup times ($st_{j_{i}j_{k}m}$), worker-machine compatibility ($o_{wm}$, documented in Table \ref{tab:example_o}), and the implied required workforce ($r_{jm}$, provided in Table \ref{tab:example_r}) for each job-machine pairing. For simplification, a uniform setup time of $1$ is assumed for all job-to-job transitions.

\begin{table}[!ht]
    \centering
    \begin{tabular}{c|c|c|c|c|c}
                 & $j_{1}$ & $j_{2}$ & $j_{3}$ & $j_{4}$ & $j_{5}$  \\
         $m_{1}$ &  2 & - & 1 & 3 & 2 \\
         $m_{2}$ &  - & 3 & 2 & 2 & 2 \\
    \end{tabular}
    \caption{Processing time ($pt_{jm}$) representation for the illustrative example. }
    \label{tab:example_pt}
\end{table}

\begin{table}[!ht]
    \centering
    \begin{tabular}{c|c|c|c|c|c}
                 & $j_{1}$ & $j_{2}$ & $j_{3}$ & $j_{4}$ & $j_{5}$  \\
         $m_{1}$ &  1 & - & 2 & 1 & 2 \\
         $m_{2}$ &  - & 1 & 1 & 2 & 1 \\
    \end{tabular}
    \caption{Required workforce ($r_{jm}$) representation for the illustrative example. }
    \label{tab:example_r}
\end{table}

This schedule is intended to minimize job completion times while concurrently optimizing setup times and accommodating the constraints associated with workforce availability and capacity. 

\begin{table}[!ht]
    \centering
    \begin{tabular}{c|c|c}
                 & $w_{1}$ & $w_{2}$   \\
         $m_{1}$ &  1 & 1 \\
         $m_{2}$ &  1 & 0 \\
    \end{tabular}
    \caption{Required workforce ($o_{jm}$) representation for the illustrative example.}
    \label{tab:example_o}
\end{table}

\begin{figure}[ht]
    \centering
    \includegraphics[width=\linewidth]{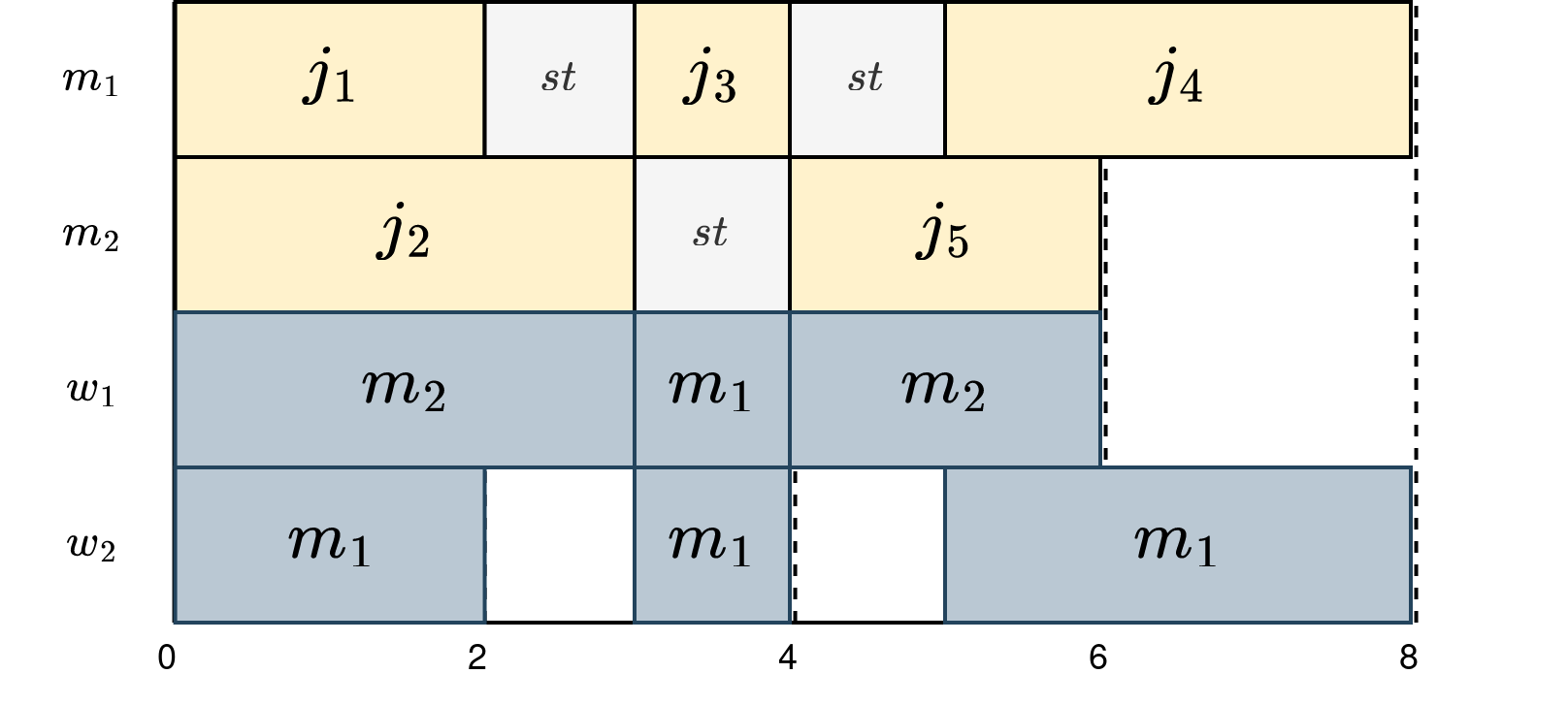}
    \caption{Illustrative example scheduled with the optimal solution: five jobs, two machines and two workers. The timeline plot is divided into machine-job (yellow) and worker-machine (grey) scheduling. }
    \label{fig: example}
\end{figure}

In Figure \ref{fig: example}, the solution completes the jobs within 8 hours. However, it is crucial to recognize alternative solutions within this particular example that might suit the scheduling necessities. Multi-objective problems have a trade-off between the individual objectives of the cost function that can modify the scheduling conduct (see equation \ref{eq: multi-objective}).

For instance, consider job $j_4$, which could be finished in 2 hours in the machine $m_2$. This is an option to reduce makespan. However, assigning $j_4$ to $m_2$ carries some implications. It would require the dedication of two workers to this particular machine, blocking their availability for other tasks. As a result, job $j_5$ would inevitably start later than initially scheduled, also causing an additional setup on machine $m_2$.

This trade-off illustrates the complex decision-making process involved in the Unrelated Parallel Machine Scheduling problem. While alternative solutions may seem appealing regarding individual task completion times, they often introduce resource allocation and setup time difficulties. Consequently, achieving the optimal balance between minimizing makespan, amount of setups, and resource utilization is a challenge within this problem domain. Exploring this example helps to understand the complexities and possible solutions for the Unrelated Parallel Machine Scheduling problem.

\section{RL environment}
\label{sec: environment}

This section introduces the essential elements of the Reinforcement Learning environment implemented in this approach. We explain fundamental concepts such as the action space, observation space, and the intricacies of the reward function. Additionally, we delve into the necessary adaptations for a Multi-Agent scenario, establishing the groundwork for a comprehensive understanding of the functionality and efficacy of this approach in complex decision-making scenarios.

In Reinforcement Learning, key concepts include the environment (a container holding information about the whole system), the agent (the entity interacting with the environment), the definition of states (describing the current situation), actions (possible moves) and rewards (numeric feedback on decision effectiveness). The agent interacts with the environment through actions, leading to state changes and the receipt of rewards based on goal achievement. These rewards may not always be directly linked to a single action but can result from a sequence of prior actions.

\begin{figure}[ht]
    \centering
    \includegraphics[width=\linewidth]{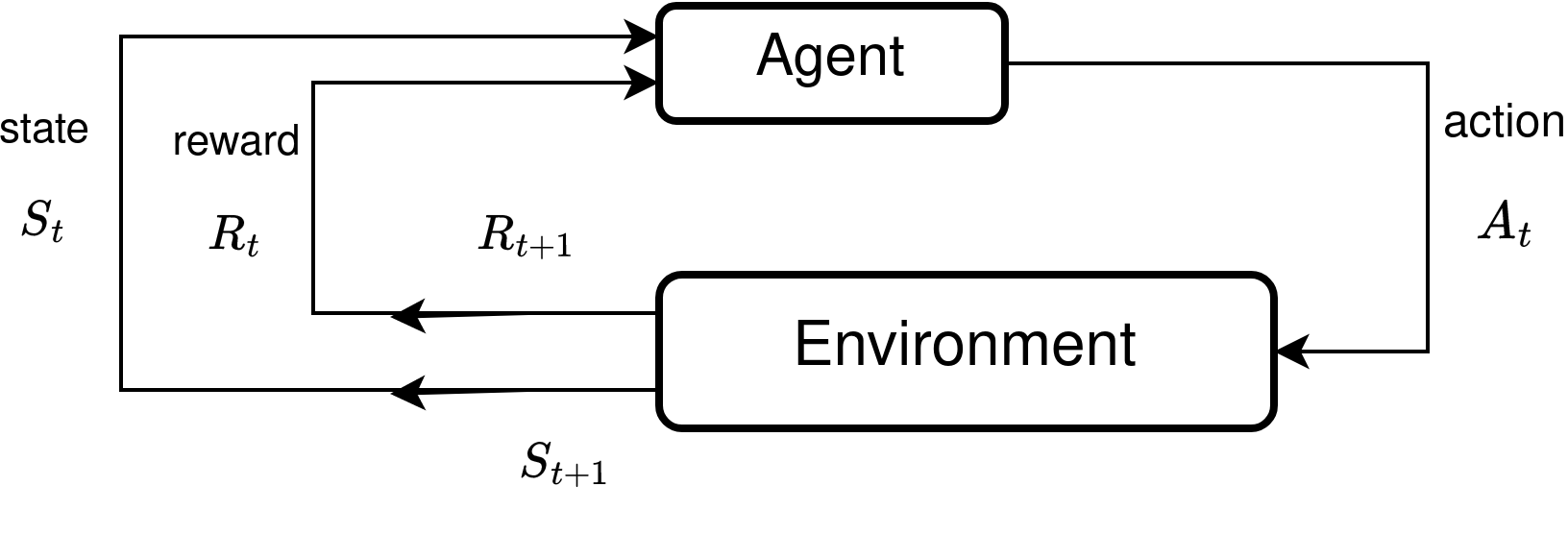}
    \caption{A simplified representation of a Reinforcement Learning System.  }
    \label{fig: example_rl}
\end{figure}

This technique implies optimizing the selection of actions to perform through policies (denoted as $\pi$), which are functions mapping states to actions,$\pi(s)$. It can also signify the probability of taking action $A_t$ in a given state $S_t$, $\pi(a|s)$. The objective is to maximize the cumulative rewards in the long term contingent upon the selected actions. This pursuit leads to seeking the optimal policy, which helps to make the optimal decision in each state.

\subsection{Single-agent approach} 
\label{sub: single}

The environment comprises plant machines, remaining jobs, and available resources. As mentioned, the jobs are characterized by processing time, setup time, and resource requirements, the variables related to the objective function.

A pivotal aspect of the Single-Agent environment is the concept of the \textit{job-slot}. A \textit{job-slot} is a collection with a fixed size of candidate jobs the agent considers at each time step, while the remaining jobs are queued in the \textit{backlog}. Except for their quantity, the agent is unaware of the specific jobs in the \textit{backlog} at each time step. The \textit{job-slot} mechanism ensures a constant number of jobs are evaluated at each time step, standardizing the observation and action space across different scenarios and reducing computational demands.

\subsubsection{Action space}
\label{subsub:act_space_sa}

In this framework, an action represents assigning a job to a machine. The action space consists of $n$ job-machine pairings, where $n$ equals the \textit{job-slot} size multiplied by the number of machines, plus one representing the \textit{do-nothing} action. When the action \textit{do-nothing} is selected by the agent, it means there is no other possible action, and the environment passes to the next time step.

This fixed-size action space is maintained throughout all time steps. The remaining slots are treated as null action if fewer jobs than the \textit{job-slot} size are available. This consistency is advantageous for training neural networks, which typically require fixed-size input and output layers, considered as a type of padding.

However, this method has a notable drawback: it can result in invalid job-machine pairings, causing some machines to be nonassignable until a compatible job enters the \textit{job-slot}. For instance, if all jobs in the \textit{job-slot} can only be processed by an already occupied machine, the other machines cannot be allocated until the machine completes its current task and a new job is assigned. When this occurs, the \textit{job-slot} is refreshed with a new job, allowing for assignments to other machines.


\begin{table*}[ht]
    \centering
    \begin{tabular}{|c|l|c|c|}
    \hline
    \textbf{Action} & \textbf{Condition} & \textbf{Reward}  & \textbf{Objective}\\
    \hline
    \textit{Do nothing} &  Viable actions available & $-0.5$ & $P(\mathbf{x})$ \\
    
    & No viable actions & $+0.5$  & $P(\mathbf{x})$  \\
    \hline
    \textit{Null Action} & Selecting an empty job slot & $-0.5$ & $P(\mathbf{x})$ \\
    \hline
    \textit{[Job, Machine]} & Action infeasible (e.g., machine occupied) & $-1$ & -\\
    
    & Action feasible & $+1$ & $P(\mathbf{x})$ \\
   
    & $Machine$ with min. \textit{total time} to perform $Job$ & $+2$ & $T(\mathbf{x})$   \\
    
    & $Machine$ with min. resources to perform $Job$ & $+2$ & $U(\mathbf{x})$  \\
    
    & Not $Machine$ with min. \textit{total time} to perform $Job$ & $-\frac{\text{difference}}{4}$ &  $T(\mathbf{x})$  \\
    
    &  Not $Machine$ with min. resources to perform $Job$ & $-\frac{\text{difference}}{4}$ & $U(\mathbf{x})$ \\
    \hline
    \end{tabular}
    \caption{Components of the Reward Function, where column \textit{objective} represents the principal objective that is optimizing in the multi-objective function \ref{eq: multi-objective}. \textit{Total time} represents the processing time of job \textit{j} in machine \textit{m} plus the setup time.}
    \label{tab:reward-components}
\end{table*}

\subsubsection{Observation space}
\label{subsub:obs_space_sa}

The observation space characterizes the necessary information for the policy to determine the optimal action. In this environment, the observation is represented as an array of essential details associated with each job-machine pairing ($j,m$) within the context of the current decision step. Specifically, five informative elements are stored for every job-machine assignment.

\begin{itemize}
    \item \textbf{Remaining processing time} is registered ($RT(m)$), providing insights into the temporal aspects of the ongoing job ($j_{m}$) and the availability of the machine. In case the machine is not working, it is zero.
    \item \textbf{Available resources} is included, contributing to assessing the human resources available to operate the machine ($AR_{m}$).
    \item \textbf{Total scheduled time} of the machine ($T(m)$) is another crucial element, influencing decisions to establish a homogeneous workload distribution among the various machines. 
    \item \textbf{Total time} required for job $j$ execution, comprising both processing time and setup time ($pt_{jm} + st_{j_{m}jm}$).
    \item \textbf{The required resources} that the job demands for its execution on the specified machine ($r_{jm}$).
\end{itemize}

Similar to the challenges encountered in defining the Action space, the Observation space presents analogous considerations. The Observation space's size correlates to the number of unique job-machine combinations under review at a given moment. Notably, the observation serves as the input for the policy, necessitating a fixed size to align with the requirements of the underlying neural network architecture.

Moreover, considering all possible job-machine pairings, a comprehensive exploration of the observation space poses a computational challenge for the agent. The combinatorial explosion of potential combinations generates an exhaustive exploration that is impractical. Hence, a strategic and efficient approach is critical. Using the \textit{job-slot} is a reasonable choice, allowing for more efficient exploration of the observation space and facilitating effective learning.

\subsubsection{Objective and Reward Design}
\label{subsub:reward_sa}

In pursuing an effective scheduling strategy, this approach aims to optimize the allocation of jobs to suitable machines while concurrently minimizing the total completion time and resource utilization across all tasks. 
Notably, the optimization process does not account for allocating specific resources to individual jobs; instead, resources are assigned using a random algorithm.

The formulation of this multi-objective function is expressed mathematically in the equation \ref{eq: multi-objective}, which considers the minimization of the total completion time (setup- and processing time), the minimization of resource utilization and the maximization of the number of assigned jobs. This objective encapsulates the core challenges of resource allocation and task completion efficiency within the given problem domain.

The reward function is designed to guide the RL agent toward learning a scheduling strategy that achieves the overarching objective and addresses scenarios that may occur during action-making. In this implementation, the reward function is cumulative. It is built by various components, each capturing a specific aspect of the decision-making process. The key insights from the reward function are presented in Table \ref{tab:reward-components}.

Collectively, these components create a reward system that encourages the RL agent to make decisions aligning with the overarching scheduling objective. Furthermore, this reward makes the model assign jobs to machines with fewer resource requirements and processing time. In this sense, total time comprises the processing and setup time, making the system select the optimal sequences and reducing setup between the machine's jobs.

During the experimentation, it became clear that the initial reward setting failed to prevent undesirable learning outcomes. Specifically, during training, the agent repeatedly chose the \textit{do-nothing} action or selected \textit{empty job-slots}. This unintended behaviour posed a challenge to achieving optimal scheduling.

Therefore, to address this issue, the reward function includes additional components to steer the agent away from non-optimal decisions and facilitate convergence. These supplementary rewards serve as corrective mechanisms to prevent persistently opting for actions that do not contribute to the scheduling objective. The introduced corrective measures are as follows:

\begin{itemize}
    \item \textbf{Excessive Inaction Penalization:}
        if the agent chooses the \textit{do-nothing} action more repeatedly when alternative actions are feasible, it triggers a proactive intervention. In response, a substantial \textit{malus} of $-10$ is applied to the reward, acting as a deterrent to the undesired excessive inaction.

    \item  \textbf{Sequential Empty Slot Avoidance:}
        if the agent repeatedly chooses an \textit{empty job-slot}, an additional \textit{malus} of $-1$ is imposed on the reward.
        This penalty breaks the sequential pattern of choosing unproductive actions, redirecting the agent towards more constructive decision-making.

\end{itemize}

By incorporating these supplementary rewards, the agent is guided away from sub-optimal strategies, promoting a more active exploration of the action space and encouraging the convergence of the scheduling process.

\subsection{Multi-agent approach}
\label{sub: multi}

In the multi-agent scenario, each machine is considered separately and essentially treated as an individual agent in the system. This approach simplifies the system by reducing the observation and action space to focus on the jobs associated with each specific machine. Consequently, in a problem involving M machines, M agents, action spaces, and observation spaces are defined.

A cooperative interaction approach is employed in this multi-agent environment to promote collaboration and coordination among these agents. This implies that the agents are designed to work towards a common objective instead of competing with each other.

\subsubsection{Action space}
\label{subsub:act_space_ma}

In this framework, since each machine is considered an independent agent, decision-making is reduced to jobs which can be executed on the machine. Therefore, the action space is defined by the jobs associated with the machine it represents. In parallel with the Single-Agent approach, the \textit{job-slot} maintains consistency across action and observation spaces. 


\subsubsection{Observation Space}

The observation space is individualized to include information relevant to the executable jobs on the respective machine. This individualized perspective is critical to improve effective decision-making. Following the environment characterization of the Single-Agent approach, every job $j$ assignment is represented by five values: Remaining processing time, available resources, total scheduled time, total processing time of job $j$, and required resources (workforce) to perform job $j$.

\subsubsection{Global Observation}

Centralization is critical in the multi-agent approach, wherein a coordinator plays an essential role by maintaining a global view that encompasses all agents. This coordinator acts as an additional agent but has the unique characteristic of having a global observation. This global observation aggregates the individual observations from all the agents. This agent's use and role depend on the algorithm employed. 

The global observation comprises collective insights from each agent, offering a perspective on the system's global state. In this sense, centralized coordination becomes valuable in scenarios where collaborative strategies are essential for achieving optimal outcomes.

\subsubsection{Objective and Reward Design}

The objective remains consistent with the Single-Agent framework, aiming to schedule all jobs with minimum completion time and resource utilization. The only difference is in resource allocation: the Single-Agent approach employs a random selection of workers (resources), whereas the Multi-Agent framework utilizes a greedy algorithm for allocation by considering the number of machines each worker can operate. Adopting a greedy assignment method mitigates resource allocation conflicts that may emerge in the Multi-Agent framework due to simultaneous job assignments across all machines at each time step.

In the Multi-agent approach, each agent has its reward. This reward is calculated using the same design presented in the Single-Agent method (see table \ref{tab:reward-components}). Additionally, in this framework, an extra reward component is considered to address possible conflicts between the agents. In a time step, the agents are unaware of the decision taken by other agents at this exact moment. Consequently, there is a possibility of two or more agents taking action over the same job. In such cases of overlap, a \textit{malus} of -1 is assigned to the rewards of the involved agents. 

This mechanism discourages conflicting or overlapping actions, promoting variety in agent decision-making and mitigating situations where multiple agents might inadvertently choose the same action.

\section{Methodology}
\label{sec: methodology}

This section presents an in-depth analysis of the algorithms employed in this research. The primary objective is to conduct a meticulous comparative assessment, commencing with the comprehensive exposition of Single-Agent methodologies. Four distinct algorithms have been implemented within this scope. Following this exploration, the analysis extends into the domain of Multi-Agent approaches, thereby significantly broadening the research landscape. This methodology section is an indispensable resource, facilitating a comprehensive understanding of these algorithmic implementations and their applicability in addressing complex decision-making scenarios.

\subsection{Single-agent}
\label{sub: single_method}

In the context of this Single-Agent framework, the work incorporated four algorithms, namely Deep Q-Learning (DQN) \citep{reyna2015reinforcement}, Advantage Actor-Critic (A2C) \citep{Volodymyr}, Proximal Policy Optimization (PPO) \citep{schulman2017proximal}, and Maskable PPO \citep{journals/corr/abs-2006-14171}. 
All these algorithms combine reinforcement learning with deep neural networks, allowing agents to make decisions based on maximizing estimated cumulative rewards.
A brief description will outline their distinctive characteristics and limitations.

\textbf{Deep Q-Learning} (DQN) is based on Q-learning, which aims to approximate the optimal action-value function (Q-function), indicating the highest expected reward an agent can achieve by executing each possible action in a specific state. In the interaction phase, the agent receives environmental observations and utilizes the Q-network to decide the most promising action, recording experiences into a buffer. In the learning phase, a batch of experiences is randomly selected to train the deep neural network, aiming to minimize the discrepancy between predicted and ground-truth Q-values. Additionally, the algorithm employs a target network to stabilize learning and an epsilon-greedy strategy to ensure the exploration of the environment \citep{wunder2010classes}. DQN has demonstrated remarkable success in complex environments, surpassing human-level performance in various Atari games \citep{van2016deep}. 

\textbf{Advantage Actor-Critic} (A2C) is an advanced variant of the Actor-Critic algorithm \citep{a2cpaper}, which leverages the Advantage function to stabilize training and reduce variance. In this approach, the actor represents the policy function, guiding the agent's actions, and the critic estimates the Q-value function. The critic and the actor functions are parameterized with neural networks. The A2C efficiently combines policy and value-based approaches, allowing faster convergence in continuous action spaces. However, A2C can encounter challenges related to elevated variance during training, potentially leading to slower convergence. Furthermore, implementing the Advantage function involves utilizing two value functions, which may increase computational complexity and memory usage. Despite these limitations, A2C remains a powerful tool for training agents in complex environments, facilitating exploring and exploiting action spaces.

\textbf{Proximal Policy Optimization} (PPO) is a reinforcement learning method that incorporates restrictions as penalties into the objective function to overcome the drawbacks of conventional policy gradient approaches. It simplifies the calculations of the TRPO algorithm \citep{schulman2017trust} and enhances optimization, allowing sporadic constraint violations by enabling first-order optimizers like the Gradient Descent approach. PPO employs a clipped surrogate objective function to limit policy updates within a trusted region, maintaining stability and reliability during optimization. This approach minimizes deviations from the initial policy and enhances the likelihood of converging towards an optimal solution. It offers a potential way to solve large-scale problems due to its versatility and effectiveness. PPO, which compromises robust optimization performance, is essential for handling complex and dynamic situations. It marks a substantial advancement in reinforcement learning algorithms.

\textbf{Maskable PPO} is a variant of the Proximal Policy Optimization (PPO) method that is extended using the technique of invalid action masking \citep{journals/corr/abs-2006-14171}. This strategy is used to prevent the execution of invalid actions while training policy gradient algorithms. In contrast to assigning negative rewards for invalid actions, avoiding selecting invalid actions by using a mask is more efficient and less prone to failure in scenarios with extensive entire action space. It is shown that this approach scales when dealing with an ample space of invalid actions, especially in complex scenarios.

\subsection{Multi-agent}
\label{sub: multi_method}

The work incorporated three algorithms in this Multi-Agent framework. A brief, comprehensive description will outline their distinctive characteristics and limitations. Counterfactual Multi-Agent Policy Gradients (COMA) and Multi-Agent PPO (MAPPO) share the same base structure: Centralized training with Decentralized execution (CTDE) composed of a centralized critic (built using the global state) and decentralized actors \citep{ikeda2022centralized}. At the same time, Multi-Agent Trust Region Policy Optimization (MATRPO) is an entirely decentralized algorithm.

\textbf{Counterfactual Multi-Agent Policy Gradients} by Foerster et al. \cite{foerster2018counterfactual} is a Multi-Agent actor-critic approach utilizing a centralized critic to train decentralized actors for each agent. It involves estimating a counterfactual advantage function for separate agents addressing Multi-Agent reward assignments. The method uses a counterfactual baseline to assess a single agent action while keeping others fixed. The centralized critic representation provides efficient computation of the counterfactual baseline. The advantage function calculates a distinct baseline using the centralized critic to determine counterfactuals when only the action of agent \textit{a} changes. The equation comprises two terms: the current selected action's global Q-value and the expectation of a global Q-value under all possible actions of agent \textit{a}.

\textbf{Multi-Agent PPO} by Yu et al. \cite{yu2022surprising} is an approach that uses the PPO algorithm for Multi-Agent implementations. This on-policy algorithm is less dominant in multi-agent scenarios than its off-policy counterparts due to the perceived lower sample efficiency in such environments. Each agent follows the standard PPO training algorithm by adding a centralized value function for calculating the  Generalized Advantage Estimation (GAE), a policy gradient estimator, and conducting the PPO critic learning procedure.
This algorithm has demonstrated strong performance across diverse Multi-Agent challenges, achieving robust results with slight tuning and lacking domain-specific modifications \citep{yu2022surprising}. MAPPO often demonstrates competitive results while maintaining comparable sample efficiency to off-policy baselines.

\textbf{Multi-Agent Trust Region Policy Optimization} by Li et al.\cite{li2023multiagent} extended Trust Region Policy Optimization (TRPO) \citep{schulman2015trust} for Multi-Agent reinforcement learning. This approach transforms the TRPO policy update into a distributed consensus optimization. This algorithm optimizes distributed policies based on local observations and individual rewards, allowing agents to operate without knowledge of global observations, rewards or policies. Agents only share local policy ratios with neighbours via a peer-to-peer communication network without the need for centralized control. This method is essential for partially observable problems.

\section{Experiments}
\label{sec: Experiments}

This section introduces the experiments undertaken during the implementation phase of the approach. The presented algorithms have distinctive characteristics, encompassing vulnerabilities and strengths. Due to these differences, defining equitable conditions for consistently training these models becomes challenging. Nevertheless, within this section, both Reinforcement Learning (RL) and Multi-Agent Reinforcement Learning (MARL) technologies undergo training within identical scenarios featuring an equivalent number of machines, jobs, and general conditions. Notably, the training duration, expressed in timesteps, stays invariant across both single- and Multi-Agent approaches to facilitate a comprehensive benchmarking of the algorithms. Moreover, a predefined hardware configuration is employed throughout the training process of deep learning models. This process aims to acquire the best-performing models, distinguishing between Single-Agent and Multi-Agent approaches. 

In the experimental phase, multiple deep neural networks were employed as the policy, for instance, Multi-Layer Perceptron \citep{gardner1998artificial}, Long Short-term Memory  \citep{hochreiter1997long} and Gated Recurrent Unit \citep{chung2014empirical}. These policies were tested for Multi-Agent approaches. However, the experiment reveals that differences in policy structure have minimal impact on results, so a simple MLP is used for the comparisons.




\subsection{Results}

In the training phase, various scenarios are generated to expose the models to diverse operational conditions. Nevertheless, the principal characteristics of the environment remain static. The models are subjected to a scenario characterized by 30 jobs (J=30), 12 machines (M=12), 60 workers (W=60), and a job-slot constraint of 10.

Notably, variables such as processing time per machine, the number of units, resource requirements, or setup times are randomly modified to introduce dynamic elements, providing realistic variations. The jobs are generated using a processing-time distribution of U(10,30) and setup time distribution of U(10,20). The resource requirement is defined using a distribution of U(1, 5) workers. This procedural approach ensures that the algorithms face scenarios with dynamic variables while maintaining static parameters, such as the number of machines and job slots, to prevent conflicts with neural network input and output sizes.

\begin{figure}[ht]
    \centering
    \includegraphics[width=\linewidth]{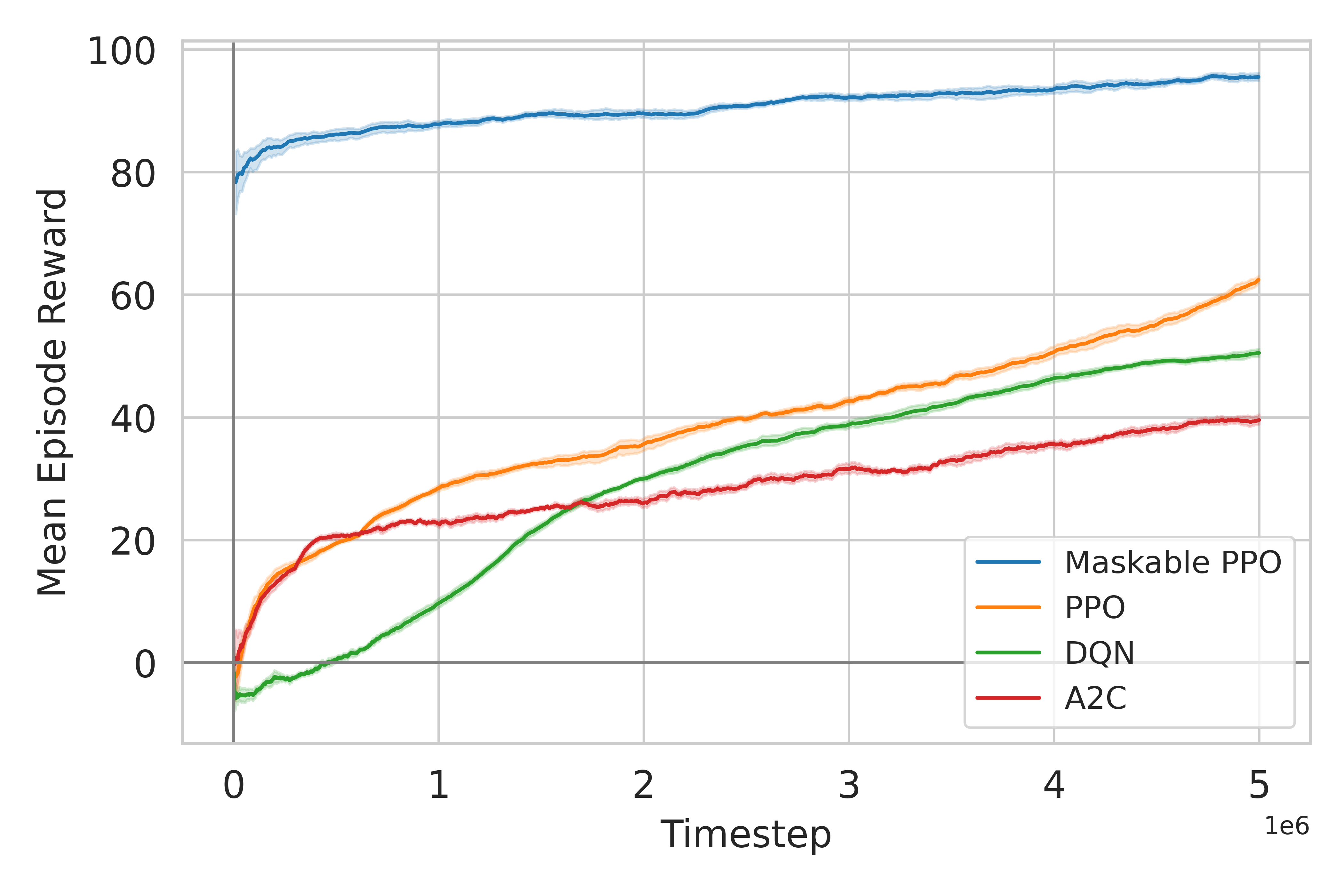}
    \caption{Training results of the four Single-Agent models during 5 million timesteps, showing the mean episode reward as a metric. The line shows the estimate of the central tendency and a confidence interval of the multiple training runs.}
    \label{fig:sarl_reward}
\end{figure}

The models undergo a training phase of five different runs of 5 million timesteps each. In this section, the episode mean reward is employed as the metric for algorithm comparison. Given that the reward remains consistent for Single- and Multi-Agent setups, this metric is appropriate for analyzing the approaches.

Figure \ref{fig:sarl_reward} illustrates the training of the Single-Agent algorithms, showing the interval of the mean episode reward of the different runs. Comparing the results obtained, it is evident that the Maskable PPO algorithm exhibits superior performance from the initial episodes. This algorithm, an extension of PPO, incorporates a valid-action mask that significantly facilitates the learning process by focusing on selecting the optimal scheduling among the possible ones, avoiding the need to deduce feasible and infeasible actions, especially in the first steps.

Notably, the results highlight that, with an increase in the number of machines or jobs, the first three algorithms fail to schedule all jobs due to termination criteria. They become stuck by repetitively choosing non-achievable actions, as indicated by the negative values accumulated in the initial steps. This signifies that the initial decisions result in negative rewards for non-achievable actions. On the other hand, the Maskable PPO model accumulates a high reward in the first episodes of the runs since it only learns from valid actions.

\begin{figure}[ht]
    \centering
    \includegraphics[width=\linewidth]{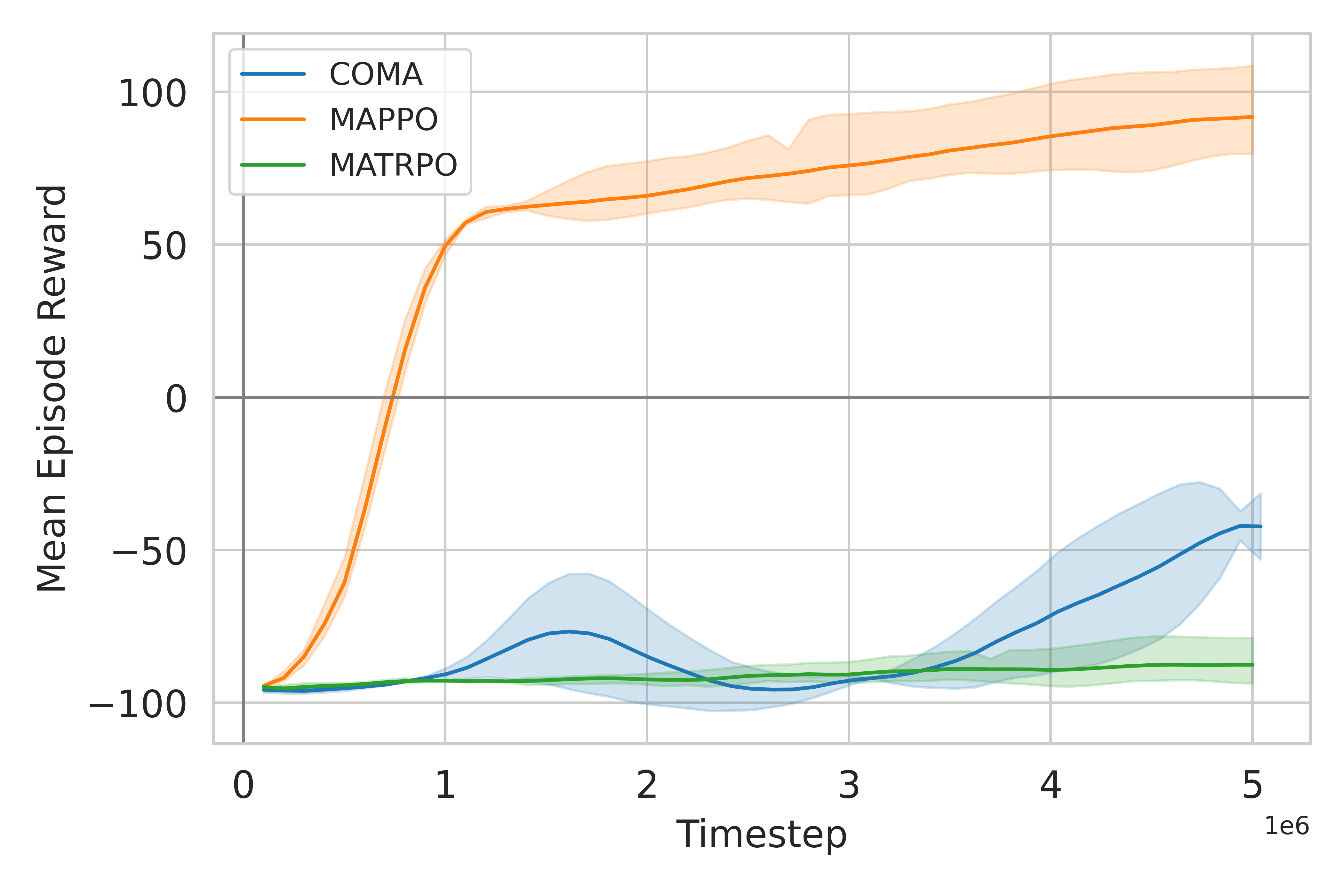}
   \caption{Training results of the Multi-agent models during 5 million timesteps, showing the mean episode reward as a metric. The line shows the estimate of the central tendency and a confidence interval of the multiple training runs.}
    \label{fig:marl_reward}
\end{figure}

The results of the Multi-Agent approach training, depicted in Figure \ref{fig:marl_reward}, clearly demonstrate that the MAPPO algorithm outperforms COMA and MATPRO. The decentralized structure of MATRPO does not allow complete information sharing among the agents, complicating the understanding of the entire system and rendering this algorithm unsuitable for this type of task. Even though COMA and MAPPO share the same base structure, MAPPO results are more stable than COMA; this might be because its counterfactual advantage suffers high variance, as some previous studies indicated \citep{DBLP, DBLP:journals/corr/abs-2006-07869}.

It is important to note that, in contrast to Single-Agent training, the MARL approach requires more time to accumulate positive rewards. This extended training duration is attributed to the intricacies of multiple agents' simultaneous learning, emphasizing the challenges in coordinating a collaborative environment.

\begin{figure}[ht]
    \centering
    \includegraphics[width=\linewidth]{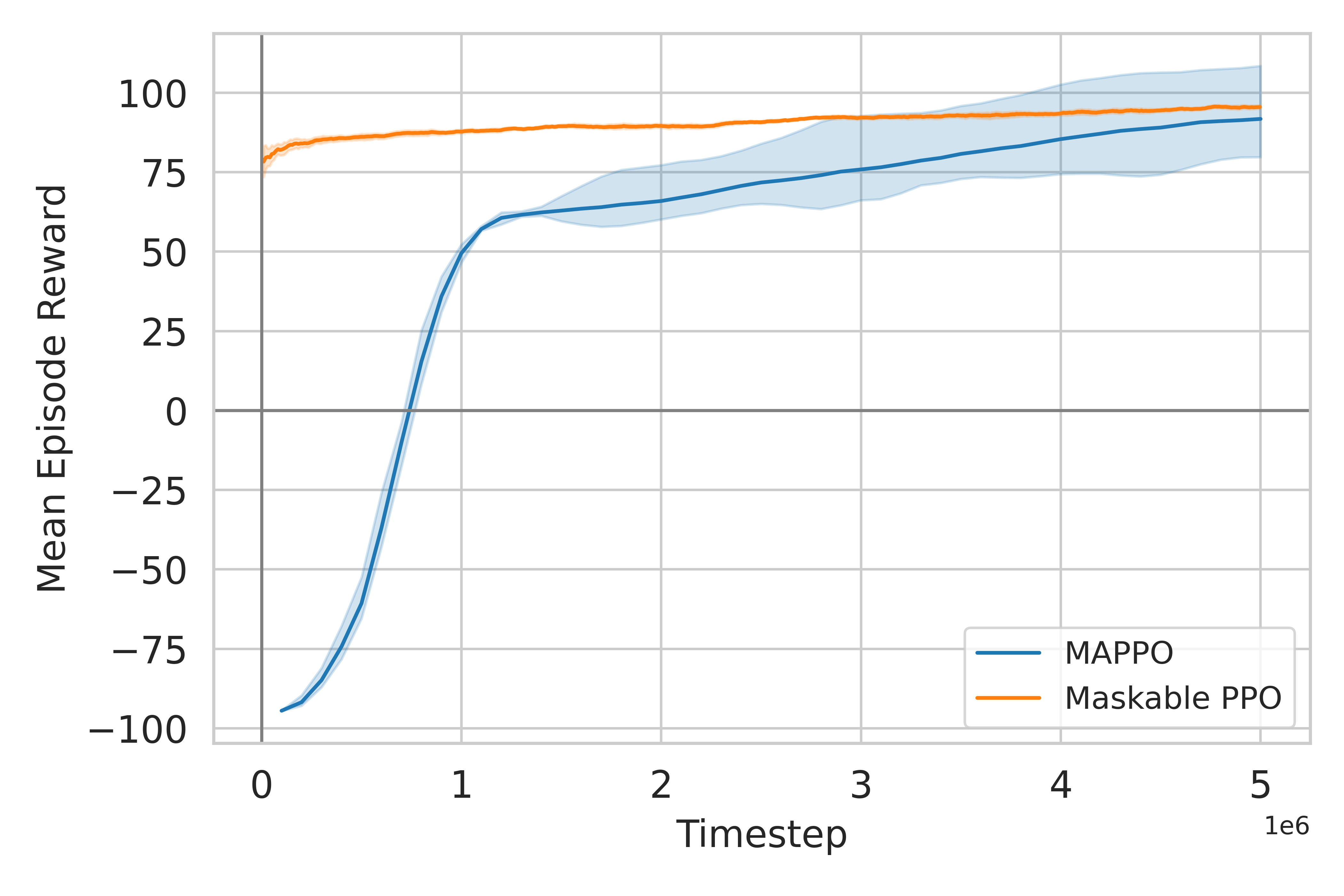}
   \caption{Training results comparison of the best Single-Agent algorithm, Maskable PPO \citep{journals/corr/abs-2006-14171}, and Multi-Agent algorithm, MAPPO \citep{yu2022surprising}. }
    \label{fig:comparisson_reward}
\end{figure}

This experimentation underscores that Single- and Multi-Agent approaches can effectively learn to schedule the problem optimally for a relatively reduced scenario, as evidenced in Figures \ref{fig:sarl_reward} and \ref{fig:marl_reward}. The Maskable PPO and MAPPO algorithms accumulate comparable rewards, close to 100 points, indicative of their decision-making capacities aligned with the scenario (see fig. \ref{fig:comparisson_reward}).
Notably, the MAPPO algorithm does not have a valid action masking, which implies it learns to distinguish between possible and impossible actions in the initial phase. Thus, the algorithm initially collects negative rewards, and then a progressive optimization of the feasible scheduling is learned.
Notwithstanding this, it successfully achieves a performance level equivalent to that of the Maskable PPO.
   
Moreover, additional reward tuning can refine the training processes to align the models with desired performance objectives. The learning curves indicate that longer training duration can improve performance, as the curve does not converge in the final time steps, leaving room for improvement. However, this validation process is enough to demonstrate insight into the performance of both approaches.

Additionally, the complexity of the learning process and the limitations within Single- and Multi-Agent algorithms are presented. While the Maskable PPO may exhibit a performance advantage over the MAPPO in the initial phase, the Multi-Agent algorithm demonstrates slightly better performance at the end of the learning process. This is crucial because these experiments were conducted with relatively few jobs and machines. The strength of MARL algorithms lies in their inherent scalability, making them adept at optimizing scheduling problems involving significantly larger numbers of jobs and machines than Single-Agent algorithms.

\section{Conclusions}
\label{sec: conclusions}

This research paper explores innovative methodologies for advanced decision-making systems, acknowledging the NP-hard nature of real-world scheduling problems. The work incorporates a Reinforcement Learning (RL) approach and presents a promising approach to address scheduling in dynamic environments with an uncertain nature. The work focuses on Multi-Agent Reinforcement Learning for the Unrelated Parallel Machine Scheduling Problem (UPMS), emphasizing its potential to adapt to non-deterministic environments.

Additionally, this article analyses the literature on Single-Agent RL algorithms, including Proximal Policy Optimization (PPO) \citep{journals/corr/abs-2006-14171}, laying the groundwork for understanding the strengths and limitations in the domain of scheduling problems (see figure \ref{fig:sarl_reward}). 


The observed results (see section \ref{sec: Experiments}) highlighted the efficacy of Single-Agent algorithms, especially of the Maskable PPO, in reduced scenarios, where training time and computational efficiency are critical. However, the scalability problems of these algorithms are evident in larger environments and action spaces. In this sense, the Multi-Agent algorithms revealed challenges in learning in a cooperative environment (see section \ref{sub: multi}), highlighting the need for further improvement in applying these algorithms to cooperative scheduling problems. However, the results obtained are promising on both Single- and Multi-Agent, specifically in both variations of PPO, the masked PPO and the Multi-Agent PPO algorithms. 

In conclusion, this research contributes by presenting the application of novel algorithms in the domain and the practical challenges and opportunities in applying MARL techniques to scheduling optimization. These findings underline the importance of balancing algorithmic sophistication with scalability to enable advancements in intelligent scheduling solutions.



\bibliography{bibliography}

\end{document}